\documentclass[runningheads]{llncs}
\usepackage[T1]{fontenc}
\usepackage{graphicx}
\usepackage{booktabs}
\usepackage{amsmath}
\usepackage{pifont}\usepackage{tikz}
\usepackage{color, colortbl}
\usepackage{authblk}
\usepackage{caption}
\usepackage{cite}
\usepackage[accsupp]{axessibility}  % Improves PDF readability for those with disabilities.
\usepackage{multirow}
\usepackage{tikz}
\usepackage{pgfplots}
\usepackage{pgfplotstable}
\usepackage{subcaption}
\usepackage{alphalph}
\usepackage{wrapfig,lipsum,booktabs}
\usepackage[symbol]{footmisc}
\definecolor{Gray}{rgb}{0.949,0.949,0.949}
\definecolor{LightCyan}{rgb}{0.88,1,1}
\definecolor{strongblue}{rgb}{0.12, 0.47, 0.71}
\definecolor{vividorange}{rgb}{1.00, 0.50, 0.05}
\definecolor{darklimegreen}{rgb}{0.17, 0.63, 0.17}
\definecolor{strongred}{rgb}{0.84, 0.15, 0.16}
\definecolor{desaturatedviolet}{rgb}{0.58, 0.40, 0.74}

\newcommand{\cmark}{\ding{51}}%
\newcommand{\xmark}{\ding{55}}%

\begin{document}
\title{CNG-SFDA: Clean-and-Noisy Region Guided Online-Offline Source-Free Domain Adaptation }
\titlerunning{CNG-SFDA}
% If the paper title is too long for the running head, you can set
% an abbreviated paper title here
%
\author{Hyeonwoo Cho\inst{1} \and
Chanmin Park\inst{2} \and
Dong-Hee Kim\inst{1} \and 
Jinyoung Kim\inst{1} \and
Won Hwa Kim\inst{3}}

\authorrunning{H. Cho et al.}
% First names are abbreviated in the running head.
% If there are more than two authors, 'et al.' is used.
%
\institute{\textsuperscript{1} VUNO Inc,  \textsuperscript{2} POSCO DX,  \textsuperscript{3} POSTECH \\ \email{hyeonwoo.cho@gmail.com}}

\maketitle              % typeset the header of the contribution
\begin{abstract}
  Domain shift occurs when training (source) and test (target) data diverge in their distribution. 
  Source-Free Domain Adaptation (SFDA) addresses this domain shift problem, aiming to adopt a trained model on the source domain to the target domain in a scenario where only a well-trained source model and unlabeled target data are available. 
  In this scenario, handling false labels in the target domain is crucial because they negatively impact the model performance. 
  To deal with this problem, we propose to update cluster prototypes (i.e., centroid of each sample cluster) 
  and their structure in the target domain formulated by the source model in online manners. 
  In the feature space, samples in different regions have different pseudo-label distribution characteristics affected by the cluster prototypes, and we adopt distinct training strategies for these samples by defining clean and noisy regions: we selectively train the target with clean pseudo-labels in the clean region, whereas we introduce mix-up inputs representing intermediate features between clean and noisy regions to increase the compactness of the cluster. We conducted extensive experiments on multiple datasets in online/offline SFDA settings, whose results demonstrate that our method, {CNG-SFDA}, achieves state-of-the-art for most cases. 
  Code is available at \url{https://github.com/hyeonwoocho7/CNG-SFDA}.
\end{abstract}
\section{Introduction}
\label{sec:intro}

Domain shift occurs when training (source) and testing (target) data diverge in their distributions, and a trained model fails to generalize on the target data.
This problem is typically dealt with Unsupervised Domain Adaptation (UDA) methods \cite{ganin2015unsupervised,long2016unsupervised, saito2018maximum,cho2022effective,wilson2020survey}, assuming that one has access to both the source and target data to align their distributions.
However, such a scenario is not feasible in many real-world cases where privacy issues or regulations 
prevent access to the source data (e.g., healthcare data or autonomous vehicles). 
Consequently, Source-Free Domain Adaptation (SFDA) \cite{liang2020we, chen2022contrastive, karim2023c, zhang2023class} has been proposed and studied, which utilizes the knowledge from a pre-trained source model and unlabeled target data.

\begin{figure}[tp]
    \centering
    \includegraphics[width=\linewidth]{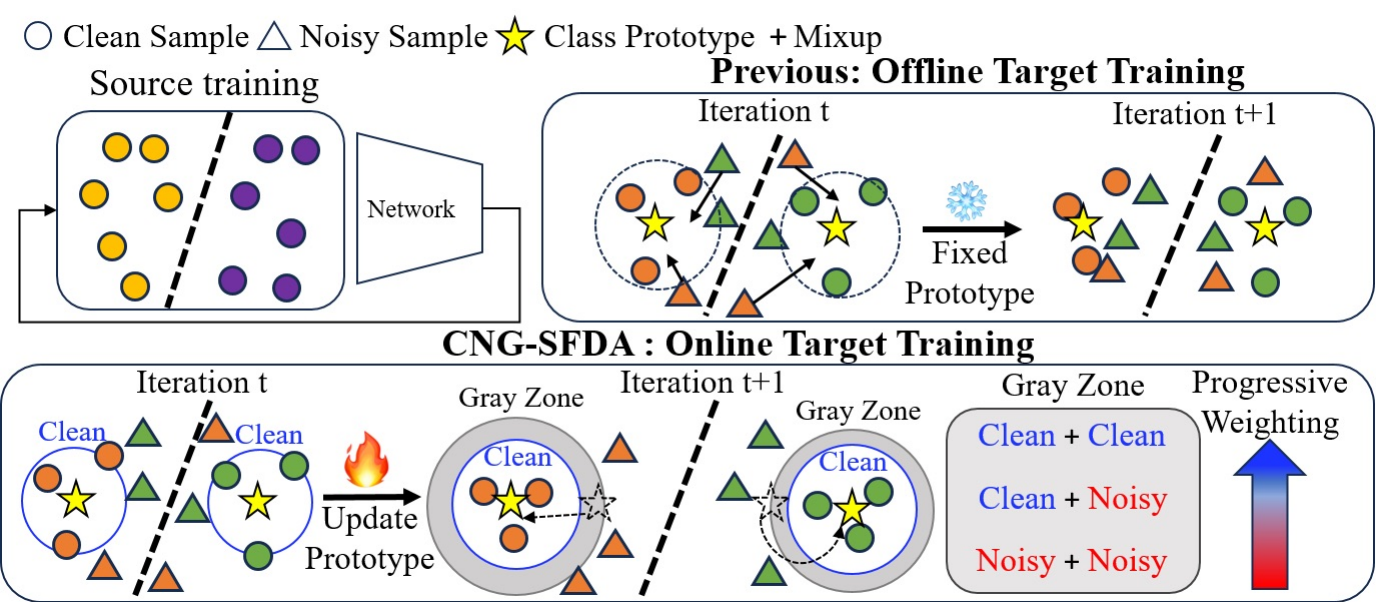}
    \caption{The upper left presents the feature distribution of clean samples on the source domain. The upper right shows the previous method that utilizes fixed class prototypes in offline manner and learn nearest neighborhoods of class prototypes. The bottom depicts the process of CNG-SFDA that updates class prototypes in online manners and learns intermediate features in gray zone with progressive weight.}
    \label{fig:Idea}
\end{figure}

Recent advancements in SFDA primarily employ self-training methods \cite{yang2021exploiting, liang2020we, chen2022contrastive, karim2023c, zhang2023class, litrico2023guiding}. 
These techniques allow a source model to learn the target domain's knowledge using pseudo-labels generated in the target domain and predominantly focus on generating reliable pseudo-labels from the neighborhood structure within the feature space. 
%rather than on how to handle reliable (clean) and unreliable (noisy) pseudo-labels produced from the target domain. 
However, due to the domain shift, there are several ambiguous and noisy samples. 
Direct utilization of their noisy pseudo-labels in training can adversely impact the model; 
hence, distinguishing and appropriately managing clean (accurate) and noisy (inaccurate) pseudo-labels is highly critical. 

Our preliminary studies, as well as other literature \cite{choi2019pseudo, zhang2021prototypical, zhang2022boostmis}, have shown that samples with noisy pseudo-labels are mostly observed at cluster boundaries
as shown in Fig. \ref{fig:Idea}, underscoring the importance of considering clean and noisy pseudo-labels separately for each cluster. 
As illustrated in Fig. \ref{fig:Idea}, several SFDA methods based self-training \cite{ding2023proxymix, du2023generation, choi2022improving, zhang2023class} introduce class prototype using
source classifier and utilize limited pseudo-labels near these prototypes for training. 
However, the prototypes in the previous approaches are fixed during target training, 
ignorantly using unreliable samples 
when target samples arrive online. 
To fully exploit the knowledge of the target domain in an SFDA scenario, 
it is crucial to update class prototypes based on target features that are added online, enabling effective learning from both reliable and unreliable pseudo-labels.
Therefore, our key idea is to design a online-offline source-free domain adaptation method that effectively learns both clean and noisy pseudo-labels in an unseen target domain, 
thereby enabling the source model to better utilize the target domain. 

In this work, we introduce Clean-and-Noisy region Guided SFDA, namely  CNG-SFDA, a novel SFDA approach that focuses on addressing pseudo-labels of samples near cluster boundaries given the domain knowledge from a source model. 
We first partition Clean and Noisy regions within clusters formed in the target domain by leveraging online-updating cluster prototype (i.e., the centroid of each cluster) to measure the distance between the prototypes and the unlabeled target samples.
Then, distinct training strategies are adopted for the clean and noisy regions due to their divergent local characteristics.

For the clean region, we conduct selective training using reliable pseudo-labels derived from predictions of the nearest neighborhoods. 
In the noisy region, we leverage mix-up inputs that represent intermediate features in gray zone bridging clean and noisy regions in Fig. \ref{fig:Idea}, unlike other methods that directly employ or discard the unreliable pseudo-labels for training. 
To draw these mix-up inputs toward the clean region effectively, we assign a ``clean probability'' to each sample indicating the likelihood of having clean pseudo-labels and introduce a mixed-clean probability defined from the clean probability of all target samples with clean and noisy, thereby enabling effective training on samples in noisy regions.
% Finally, in order to generate reliable pseudo-labels, we introduce the contrastive learning framework, aligning the nearest features effectively. 
% Our method utilizes instance-discriminate and class-semantic features by introducing the memory bank \cite{he2020momentum} and the cluster prototype embedding through contrastive learning.

To summarise, the main contributions of our work are: 
\begin{itemize}
  \item 
  We propose CNG-SFDA, a novel SFDA approach that progressively update label prototypes in the target domain for better pseudo-supervision. 
  \item
  We propose distinct training strategies for test samples with discriminate characteristics in their pseudo-label distributions. 
  \item
  We validate CNG-SFDA on various benchmarks for online/offline SFDA in single/multi-source settings, which empirically demonstrate the superiority of our approach to SOTA by a large margin in the challenging scenarios.  
\end{itemize}

\section{Related Work}
\label{sec:intro}

\subsection{Domain Adaptation Scenarios}
\label{sec:UDA-scenarios}
Table. \ref{tab:DA_settings} introduces domain adaptation (DA) scenarios addressed by CNG-SFDA. 

\noindent {\bf Domain Generalization (DG).} DG ~\cite{gulrajani2020search} aims to build a model that generalizes well to unseen target domains.
In this protocol, most approaches proposed to learn domain-invariant representations using multiple source domains.
Recently, the more challenging Source-Free Domain Generalization (SFDG) ~\cite{frikha2023towards} scenario has been handled which does not require access to the source domain. % , unlike DG.
In the end, SFDG methods aim to build a general source model from models trained on each of the multiple source domains without access to the target domain.

\begin{table*}[!t]
\centering
\resizebox{\textwidth}{!}{%
\begin{tabular}{l c c c c}
\toprule
\multirow{2}{*}{Settings} & \multicolumn{2}{c}{Source Domains} & \multicolumn{2}{c}{Target Domains Access} \\
\cline{2-5}
& Source-Free & Multi-Source & Offline & Online \\
\midrule
Domain Generalization (DG) & \xmark & \cmark & \xmark & \xmark \\
Source-Free Domain Generalization (SFDG) & \cmark  & \cmark & \xmark & \xmark \\
Unsupervised Domain Adaptation (UDA) & \xmark & \xmark & \cmark & \xmark\\
\rowcolor{Gray}Source-Free Unsupervised Domain Adaptation (SFUDA) &  \cmark & \xmark & \cmark & \xmark \\
\rowcolor{Gray}Source-Free Multi-Source Unsupervised Domain Adaptation (SFMSUDA) &  \cmark &  \cmark & \cmark & \xmark \\
\rowcolor{Gray}Test-Time adaptation (TTA) & \cmark & \xmark & \xmark & \cmark \\
\bottomrule
\end{tabular}}
\caption{Domain Adaptation Scenarios. We address the highlighted rows.}
\label{tab:DA_settings}
\end{table*}

\noindent {\bf Unsupervised Domain Adaptation (UDA).} %the third row of Table. \ref{tab:DA_settings}, 
The main task for UDA ~\cite{ganin2015unsupervised, long2016unsupervised} is to align the source distribution and target distribution from labeled source domain and unlabeled target domain. UDA leverages adversarial learning ~\cite{ganin2015unsupervised, tzeng2017adversarial, long2018conditional, zhang2018collaborative, cui2020gradually, chen2022reusing} and generative approaches ~\cite{bousmalis2017unsupervised, murez2018image, nam2021reducing} to reduce the discrepancy between the distributions of source and target domains.

\noindent {\bf Our Approach.}
As shown in Table \ref{tab:DA_settings}, CNG-SFDA addresses Source-Free Unsupervised Domain Adaptation (SFUDA) \cite{liang2020we}, Source-Free Multi-Source Unsupervised Domain Adaptation (SFMSUDA) \cite{ahmed2021unsupervised}, and Test-Time adaptation (TTA)\cite{wang2020tent}. The goal of SFUDA is to adapt a trained model on the source domain to the target domain where the pre-trained source model and unlabeled target data are available. During adaptation, SFUDA has offline access to the target domain. In other words, the pre-trained source model learns the target domain's knowledge by training on the target domain for several epochs. On the other hand, SFMSUDA aims to adapt a trained model on multiple source domains to an unseen target domain. In this protocol, as with SFUDA and SFMSUDA, access to the target domain is possible offline during adaptation. Finally, TTA is defined as a problem of accessing target domains online, unlike previous scenarios. Here, `online' means that the pre-trained source model is accessible to the target domain during inference; hence we train only 1 epoch during adaptation's phase.

In this work, we define SFUDA and SFMSUDA as offline source-freee domain adaptation (offline SFDA) because they are takes place offline. Then,  TTA is defined as online source-free domain adaptation (online SFDA).

\subsection{Online-Offline Source-Free Domain Adaptation}
\label{sec:SFUDA}
Recently, several methods \cite{liang2020we, wang2020tent, chen2022contrastive, karim2023c, zhang2023class, litrico2023guiding} have been proposed for handling the domain shift in online-offline source-free domain adaptation (SFDA) settings. SHOT \cite{liang2020we} introduces a class centroid used for pseudo-labeling for offline SFDA. Tent \cite{wang2020tent} optimizes entropy minimization for online SFDA. CoWA-JMDS \cite{lee2022confidence} utilizes the joint model-data structure as sample-wise weights to represent target domain knowledge effectively. CRCo \cite{zhang2023class} proposes a probability-based similarity between target samples by embedding the source domain class relationship.

In addition, self-training methods \cite{chen2022contrastive, ding2023proxymix, du2023generation, karim2023c} have been proposed as well. In particular, ProxyMix and PS \cite{ding2023proxymix, du2023generation} conceptualize the target domain as disjoint parts, such as proxy source and target domain. %Our method is also aligned with this approach. 
PS \cite{du2023generation} gathers samples with low entropy and incorporates them with mix-up regularization for training. ProxyMix \cite{ding2023proxymix} defines the weights of the pretrained source model as class prototypes and assumes nearby target samples as the proxy source domain. Then, it applies inter-domain mix-up between the proxy source domain and the remaining target samples, as well as intra-domain mix-up among target samples, to support the linear behavior among training samples. 

\noindent
{\bf Limitation.} The aforementioned methods selectively train only reliable samples or neglect to consider the characteristics of both reliable and unreliable samples during mix-up regularization. Additionally, since the class prototype is not updated during training, it may be challenging to apply in online environments where the domain changes.

\noindent
{\bf Our approach.} To address these limitations, we propose a method suitable for both offline and online Source-Free Domain Adaptation (SFDA) by defining class prototypes and clean-and-noisy regions that are updated online, and assigning the ``clean probability'' to each sample indicating the likelihood of having clean pseudo-labels. In particular, our method enhances the linear behavior for target samples by adjusting the model's mix-up weight using the mixed-clean probability, defined from the clean probability of all target samples with clean and noisy pseudo-labels, during mix-up regularization.
% Especially, by adjusting the model's mix-up weight differently based on the clean probability of all target samples, having clean and noisy pseudo-labels, our method enhances the linear behavior among target samples. 

\begin{figure*}[tp]
    \centering
    \includegraphics[width=\textwidth]{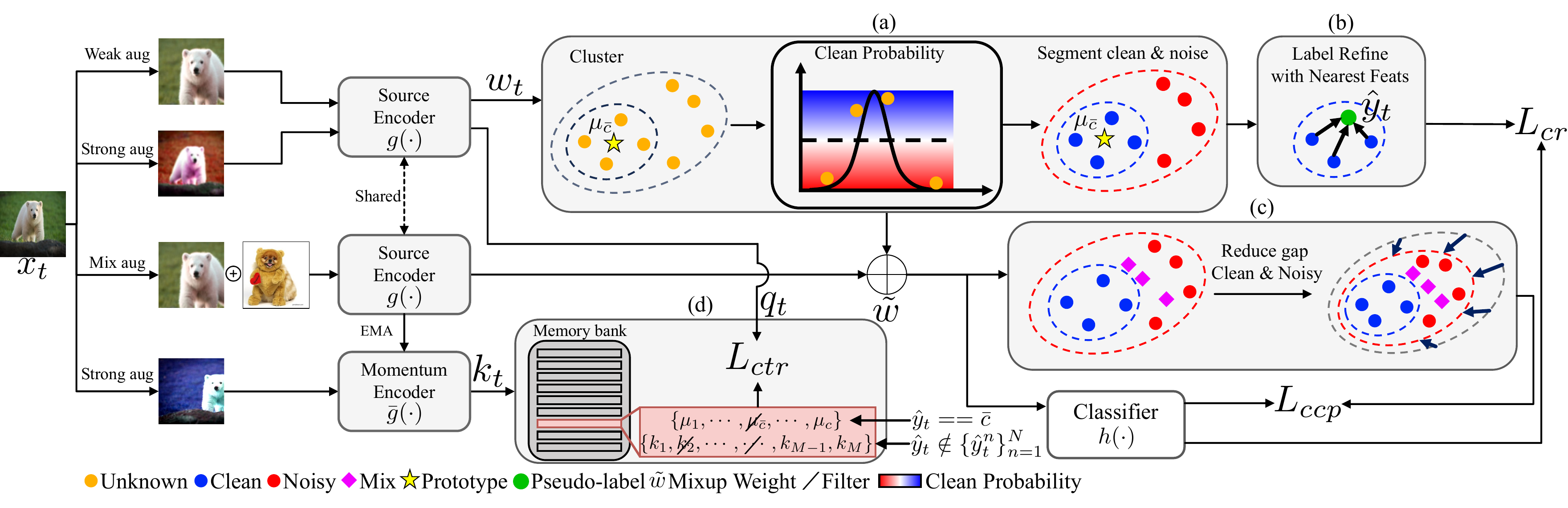}
    \caption{\textbf{Framework of CNG-SFDA} (a): partition the cluster into `Clean' (close to the cluster prototype) and `Noisy' (far from the cluster prototype) region based on the clean probability distribution. (b): training samples in clean regions with pseudo-labels generated from prediction of closest features. (c): reducing a gap between clean and noisy region with mix-up weight. (d): prototype/instance-aware contrastive learning.}
    \label{fig:Overview}
\end{figure*}
\section{The Proposed Method}
\label{sec:method}
 In this work, we address Source-Free Domain Adaptation (SFDA) for the image classification task. In SFDA, we are given a source model trained on labeled source data $\mathcal{D}_s=\{x_s^i, y_s^i\}_{i=1}^{N_s}$ and unlabeled target data $\mathcal{D}_t=\{x_t^i\}_{i=1}^{N_t}$, where $x_s^i$, $x_t^i$ represent the input images and $N_s$, $N_t$ denote the numbers of data in the source and target domains, respectively. We focus on the closed-set SFDA where the target domain shares the same $C$ classes as the source domain. 
 
 Given a pre-trained source model $f(\cdot)$ only, the objective of SFDA is to adapt the model to work on the unlabeled target data $\mathcal{D}_t$. We assume a general architecture of the source model $f(\cdot)=h(g(\cdot))$ comprising an encoder $g(\cdot)$ and a classifier $h(\cdot)$, and our method is introduced below. 

 \noindent
\textbf{Overview.}
In the domain adaptation phase, we employ MoCo framework following \cite{chen2022contrastive}. 
We initialize the momentum model $\bar{f}(\cdot)=\bar{h}(\bar{g}(\cdot))$ with the parameters from the source model $f(\cdot)$ at the beginning of adaptation, and $\bar{f}(\cdot)$ is updated by the exponential moving average of the source encoder $g(\cdot)$ and the source classifier $h(\cdot)$ at each mini-batch step to enhance the stability of the feature space. As illustrated in Fig. \ref{fig:Overview}, the source encoder $g(\cdot)$ and the momentum encoder $\bar{g}(\cdot)$ take input images $\{\mathcal{T}(x_t^i)\}_{i=1}^{N_t}$ from the target domain $\{x_t^i\}_{i=1}^{N_t}$ with weak and two strong augmentations $\mathcal{T}=\{T_{w},T_{s},T_{\bar{s}}\}$ and mixed images $\mathcal{D}_{mix}=\{\tilde{x}_t^i\}_{i=1}^{N_t}$ generated from $\{x_t^i\}_{i=1}^{N_t}$, respectively. 

Then, for the purpose of categorizing the target data into individual clusters, we produce pseudo-labels $\{\hat{y}_{t}^{i}\}_{i=1}^{N_t}$ for the target data via a nearest neighbor soft voting \cite{mitchell2001soft} in a memory queue $Q_{w}$, representing the target feature space. 
As depicted in Fig \ref{fig:Overview} (a), to filter the target samples with noisy pseudo-labels $\hat{y}_{t}$ in each cluster, 
we partition the %target domain's 
clusters into clean (close to $\mu_k$) and noisy (far from $\mu_k$) regions using a clean probability derived via the distance from the cluster prototype $\mu_k$ (i.e., centroid for $k$-th cluster).

Next, to leverage different characteristics of the clean and noisy regions, we employ distinct training strategies for each region. For samples in the clean regions, we train $g(\cdot)$ and $h(\cdot)$ on them with the clean pseudo-labels $\hat{y}_{t}$ generated through nearest neighbors features, as shown in Fig. \ref{fig:Overview} (b). In contrast, for those in the noisy regions, we employ a cluster compactness learning strategy that incorporates mix-up \cite{zhang2017mixup} between samples and the clean probability of each sample, as shown in (c) of Fig. \ref{fig:Overview}. 

Finally, as shown in Fig. \ref{fig:Overview} (d), to generate reliable $\hat{y}_{t}$ from the nearest features, we adopt contrastive learning both between the prototype and sample embeddings, as well as among the samples themselves. 

\subsection{Distinguishing Clean and Noisy Regions in Clusters}
\label{sec:section-a}
In conventional Domain Adaptation (DA) scenarios, a significant domain discrepancy between the source and target domains often results in inaccurate pseudo-labels generated via the source model, negatively impacting the model's performance on the target domain. 
In particular, SFDA is more challenging compared to traditional DA, as it offers no access to the source data, and thus efficiently leveraging the knowledge of the target domain is critical. 
In this work, we utilize the cluster structure %(i.e., clean and noisy regions within clusters)
and cluster prototype $\mu_k$ in the target as the knowledge of the target domain.

To address this, we propose an approach that guides the model to learn the cluster structure in the feature space with respect to the prototypes of each cluster in the target domain. 
Initially, to enable efficient nearest-neighbor search for generating the pseudo-labels $\hat{y}_{t}$ from the cluster structures, 
we store the features $w_{t} = g(T_{w}(x_{t}))$ and their class-wise probability $p_{t} = \sigma(f(T_{w}(x_{t})))$ in a memory queue $Q_{w}=\{w_t^{j}, p_{t}^j\}_{j=1}^M$ of the length of $M$, where $\sigma$ denotes the softmax function.
Then, in order to assign the target images ${x}_{t}$ into each cluster, we generate its pseudo-labels $\hat{y}_{t}$ by performing a soft voting \cite{mitchell2001soft} among the $K$ closest neighbors of ${x}_{t}$ in $Q_{w}$. The soft voting is done by averaging their probability outputs as: 
\begin{equation}
\small
\label{eqation:1}
\hat{p}_{t}^{(i,c)}=\frac{1}{K}\sum_{j=1}^{K}p_{t}^{(j,c)},
\end{equation} 
and $\hat{y}_{t}^{i}$ is categorized into cluster $c$ by using ${argmax_{c}} \ \hat{p}_{t}^{(i,c)}$.

To determine whether a given target sample $x_{t}^{i}$ is clean or noisy within the cluster, as shown (a) in Fig. \ref{fig:Overview}, we introduce a Gaussian Mixture Model (GMM) \cite{reynolds2009gaussian}. 
%Following \cite{reynolds2009gaussian}, 
Given the observation $x_{t}^{i}$, a probability of the latent variable $z_{i}$ that $x_{t}^i$ belongs to $k$-th cluster is defined as:
\begin{equation}
\small
    \label{eqation:3}
    \gamma_{ik}= p(z_{i}=k|x_{t}^i)  = \frac{\pi_{k}\mathcal{N}(x_{t}^i|\mu_k, \sigma_k)}{\sum_{k}\pi_{k}\mathcal{N}(x_{t}^i|\mu_k, \sigma_k)}
\end{equation}
where $z_{i}$ represents the assignment of each cluster, $\mu_{k}$ and $\sigma_{k}$ indicate the mean and variance of the Gaussian distribution $\mathcal{N}(x_{t}^i|\mu_k, \sigma_k)$ for the $k$-th cluster, and $\pi_{k}$ is the prior probability of cluster $k$, i.e., $p(z_{i}=k)$. Here, assuming that $z_{i}$ has uniform distribution, the Eq. \eqref{eqation:3} is expressed as:
\begin{equation}
\small
    \label{eqation:4}
    \begin{split}
     \gamma_{ik}=p(z_{i}=k|x_{t}^i)
     &= \frac{\mathcal{N}(x_{t}^i|\mu_k, \sigma_k)}{\sum_{k}\mathcal{N}(x_{t}^i|\mu_k, \sigma_k)}, 
    \end{split}
\end{equation}
whose $\mu_{k}$, $\sigma_k$, and $z_{i}$ can be solved via EM algorithm \cite{reynolds2009gaussian}. 

In this work, we hypothesize that $x_{t}^{i}$ has a clean label if $\hat{y}_{t}^{i}$ from our model and $z_{i}$ estimated via the GMM are identical. To connect $\hat{y}_{t}^{i}$ and $z_{i}$ in GMM, we adopt the $\mu_{k}$, $\sigma_k$, and $z_{i}$ computed using $f(\cdot)$. 
Specifically, we represent our predictions $p(\hat{y}_{t}^{i}=k|x_{t}^i,\theta)$ as the posterior probability, 
along with the output feature $g(x_{t}^i)$ as the given observation. 
Thus, following \cite{reynolds2009gaussian}, ${\mu}_{k}$, ${\sigma}_{k}$ are represented as: 
\begin{equation}
\small
\label{eqation:5}
    {\mu}_{k} = \text{norm}(\frac{\sum_{i}p(\hat{y}_{t}^{i}=k|x_{t}^i,\theta)g(x_{t}^i)}{\sum_{i}p(\hat{y}_{t}^{i}=k|x_{t}^i,\theta)}), 
    \end{equation}
\begin{equation}
\small
    {\sigma}_{k} =\frac{\sum_{i}p(\hat{y}_{t}^{i}=k|x_{t}^i,\theta)(g(x_{t}^i)-\mu_{k})^{T}(g(x_{t}^i)-\mu_{k})}{\sum_{i}p(\hat{y}_{t}^{i}=k|x_{t}^i,\theta)} 
\end{equation}
where $\text{norm}(\cdot)$ is l2-normalization such that $\left\|{\mu}_{k}\right\|_{2}$$=$$1$.
With ${\mu}_{k}, {\sigma}_{k}$ estimated from $f(\cdot)$, the Eq. (\ref{eqation:4}) is defined by the probability density function:
\begin{equation}
\small
\label{eqation:6}
\begin{split}
\gamma_{ik}&=\frac{\exp(-(g(x_{t}^i)-\mu_{k})^T(g(x_{t}^i)-\mu_{k})/2\sigma_{k})}{\sum_{k}\exp(-(g(x_{t}^i)-\mu_{k})^T(g(x_{t}^i)-\mu_{k})/2\sigma_{k})} \\
&=\frac{\exp(g(x_{t}^i)^T\mu_{k}/\sigma_{k})}{\sum_{k}\exp(g(x_{t}^i)^T\mu_{k}/\sigma_{k})}.
\end{split}
\end{equation}
Thus, the probability of the sample being clean is defined as:
\begin{equation}
\small
\label{eqation:7}
\gamma_{\hat{y}_{t}^{i}=z_{i}}=p(\hat{y}_{t}^{i}=z_{i}|x_{t}^i)=\frac{\exp(g(x_{t}^i)^T\mu_{z_{i}}/\sigma_{z_{i}})}{\sum_{k}\exp(g(x_{t}^i)^T\mu_{k}/\sigma_{k})}
\end{equation}
by using $g(x_{t}^i)^T\mu_{k}$ as the distance metric between the prototype and the sample embedding.
From this probability distribution, our method enables the partition of each cluster into closer (Clean) and farther (Noisy) regions relative to the prototype through the probability threshold $\alpha$. Thus, the samples in the clean and noisy region are represented as:
\begin{equation}
\small
\label{eqation:8}
    \{x_t^i\}_{i=1}^{N_t}= 
    \begin{cases}
    \{x_{t}^{cr}\}_{cr=1}^{N_{cr}}, & \text{if} \ p(\hat{y}_{t}^{i}=z_{i}|x_{t}^i) \geq \alpha\\
    \{x_{t}^{nr}\}_{nr=1}^{N_{nr}}, & \text{otherwise},
    \end{cases}
\end{equation}
where $N_{cr}$ and $N_{nr}$ denote the number of samples in clean and noisy regions.

%\subsection{Train Clean and Noisy region}
\subsection{Training Strategy for Clean and Noisy Regions}
\label{sec:section-b,c}
In preliminary study, we observed that clean and noisy regions within a cluster have divergent local characteristics. In the clean region, the samples, as well as their closest features, have reliable pseudo-labels $\hat{y}_{t}$ generated by Eq. \ref{eqation:1}, as they are also in the clean regions. On the other hand, the noisy regions contain samples with incorrect $\hat{y}_{t}$, as they are located near the cluster boundaries.
To deal with this, we employ distinct training strategies for $\{x_{t}^{cr}\}_{cr=1}^{N_{cr}}$ and $\{x_{t}^{nr}\}_{nr=1}^{N_{nr}}$. 

\noindent{\bf Learning from clean samples.} In the clean region, we exclusively train on $\{x_{t}^{cr}\}_{cr=1}^{N_{cr}}$ with the pseudo-labels $\hat{y}_{t}$, as they are more reliable than %$\hat{y}_{t}$ for 
$x_{t}^{nr}$ from the noisy regions. 
%The clean loss is given by:
The loss for the samples in the clean region is given by:
\begin{equation}
\small
\label{eqation:9}
    L_{cr}=-\mathbb{E}_{x_{t}^{cr} \in \mathcal{X}_{t}}\sum_{c=1}^{C}\hat{y}_{t}^{c} \ log \ p_{t}^{c}.
\end{equation} 

\noindent {\bf Learning from noisy samples.} For the noisy region, we avoid directly using $\hat{y}_{t}$ for $\{x_{t}^{nr}\}_{nr=1}^{N_{nr}}$ for training but rather leverage {\em mix-up} input $\tilde{x}_t$ for the effective noisy training, as the mix-up samples lie in between the stable and noisy samples \cite{na2021fixbi}. 
%Inspired by \cite{na2021fixbi}, 
We consider $g(\tilde{x}_t)$ as the features to bridge the gap between the clean and noisy regions. 
Here, $\tilde{x}_t^{i}$ and its pseudo-label $\tilde{y}_{t}^{i}$ are represented as $\lambda x_{t}^i+(1-\lambda) x_{t}^j$ and $\lambda \hat{y}_{t}^i+(1-\lambda) \hat{y}_{t}^j$, where $\{x_{t}^i, x_{t}^j\} \in D_{t}$ and mix-up ratio $\lambda \in \left [ 0,1 \right ]$.
To effectively attract $\tilde{x}_t$ to the clean region, we introduce the mixed-clean probability $p(\tilde{y}_{t}^{i}=z_{i}|\tilde{x}_t^{i})$, assigning greater significance as $\tilde{x}_t^{i}$ approaches the clean region. We define $p(\tilde{y}_{t}^{i}=z_{i}|\tilde{x}_t^{i})$ as a weighted sum of the clean probabilities of $x_{t}^i$ and $x_{t}^j$:
\begin{equation} 
\small
    \label{eqation:12}
    \begin{split}
p(\tilde{y}_{t}^{i}=z_{i}|\tilde{x}_t^{i})=\lambda p(z_{i} \vert x_{t}^{i}) +(1-\lambda)p(z_{j} \vert x_{t}^{j}).
    \end{split}
\end{equation}
We then represent the mix-up weight, denoted as $\tilde{w}$, using $p(\tilde{y}_{t}^{i}=z_{i}|\tilde{x}_t^{i})$ with an exponential function:
 \begin{equation}
 \small
    \label{eqation:13}
    \tilde{w}=\exp(p(\tilde{y}_{t}^{i}=z_{i}|\tilde{x}_t^{i})) 
\end{equation}
which varies depending on the regions where $x_{t}^i$ and $x_{t}^j$ are located by Eq. \ref{eqation:12}. If both $x_{t}^i$ and $x_{t}^j$ are in the noisy region, $\tilde{w}$ has a low value and vice versa for the clean region. We argue that using $\tilde{x}_t$ with $\tilde{w}$ enhances the compactness of the cluster since the model can learn intermediate features between the clean and noisy region, as illustrated in (c) in Fig. \ref{fig:Overview}. 
Together with the mix-up samples, the Cluster ComPactness (CCP) loss, denoted as $L_{ccp}$, is given by:
  \begin{equation}
  \small
    \label{eqation:14}
    L_{ccp}=-\tilde{w}\cdot\mathbb{E}_{\tilde{x}_{t}\in \mathcal{\tilde{X}}_{t}}\sum_{c=1}^{C}\tilde{y}_{t}^{c} \ log \  \tilde{p}_{t}^{c},
\end{equation}
where $\tilde{p}_{t}=\sigma(f(\tilde{x}_t))$ are the predicted probabilities for $\tilde{x}_t$. 

Plus, we add a regularization promoting diverse predictions similar to \cite{chen2022contrastive}:
\begin{equation}
\small
    \label{eqation:15}
    \begin{split}
    L_{div} = \mathbb{E}_{x_{t} \in \mathcal{X}_{t}}\sum_{c=1}^{C}\bar{p}_{t}^{c} \ log \ \bar{p}_{t}^{c} 
    \end{split}
\end{equation}
where $\bar{p}_{t}=\mathbb{E}_{x_{t} \in \mathcal{X}_{t}}\sigma(f(T_{s}(x_t)))$ and $L_{div}$ prevents the model from trusting false pseudo-labels.

\subsection{Instance and Prototype-aware Contrastive Learning}
\label{sec:section-d}
In our method, we generate the pseudo-labels of the target samples from the nearest features, as discussed in Eq. \ref{eqation:1}. However, the pseudo-labels might be incorrect if the predictions of their nearest features are not correlated with the class of the target. To better align the nearest features, we introduce the contrast learning framework.
Prior approaches \cite{oord2018representation, zhang2022divide, huynh2022boosting, kim2020mixco} consider each pair of instances as attractive and dispersing samples to enable instance discrimination.
However, the limitation of these approaches is that it is difficult to attract different instances with similar semantic features in the same cluster, 
thereby posing challenges in encoding class-semantic information between instances.

To address these challenges, 
we contrast the prototype ${\mu}_{k}$ obtained in Section \ref{sec:section-a} and queries $q_t=g(T_{s}(x_t))$ for prototype-aware contrastive learning. 
As described in Fig. \ref{fig:Overview} (d), 
we pair $q_t$ and ${\mu}_{k=\bar{c}}$ (i.e., the prototype feature of corresponding class $\bar{c}=\hat{y}_{t}$ of $q_t$), as the positive pair. 
In contrast, the negative pair is given by $q_t$ and $\{{\mu}_{k \neq \bar{c}}\}_{k=1}^C$, which are the prototype features not corresponding to class $\bar{c}$. 
Our prototype-aware contrastive loss is defined as:
\begin{equation}
\small
    \label{eqation:17}
    L_{prt}=-\log\frac{\exp\ q_{t} \cdot {\mu}_{k=\bar{c}}/\tau}{\sum_{k \neq \bar{c}} q_{t} \cdot {\mu}_{k}/\tau},
\end{equation}
and $\tau$ is a temperature controlling the scale of predictions.

Moreover, to learn instance-discriminate features, we leverage queries $q_t$ and keys $k_{t}=\bar{g}(T_{\bar{s}}(x_t))$ as the positive pair as in \cite{he2020momentum}. 
To build the negative pairs $\mathcal{N}_p$ of $q_t$, we store $k_t$ to the memory queue $Q_{k}=\{k_t^{i}\}_{i=1}^M$ with the length $M$, 
but $Q_{k}$ may have features which have the same class as the corresponding class $\hat{y}_t$ of $q_{t}$. 
In our work, to eliminate negative pairs with the same class, we utilize the nearest feature information in $Q_{k}$. 
We exclude the negative pairs when $\{\hat{y}_{t}^{n}\}_{n=1}^N$ of their $N$ closest features are at least one the same as $\hat{y}_t$, as shown in Fig. \ref{fig:Overview} (d). 
Our instance-aware contrastive loss reflecting the behavior above is given by:
 \begin{equation}
 \small
    \label{eqation:18}
    L_{inst}=-\log\frac{\exp\ q_t \cdot k_t /\tau}{\sum_{j \in \mathcal{N}_p} q_{t} \cdot k_{t}^{j}/\tau} 
\end{equation}
where $\mathcal{N}_p$ is defined as $\{\ j|1 \le j \le M, \hat{y}_{t}^j \notin \{\hat{y}_{t}^{n}\}_{n=1}^N \}\ $, and 
the overall contrastive loss combining $L_{prt}$ and $L_{inst}$ is expressed as:
\begin{equation}
\small
    \label{eqation:22}
    L_{ctr} = L_{inst} + L_{prt}.
\end{equation}

In the end, by summarizing all the losses from \eqref{eqation:9}, \eqref{eqation:14}, \eqref{eqation:15} and \eqref{eqation:22}, we get the global loss function as below:
\begin{equation}
\small
    \label{eqation:23}
    L_{overall} = \gamma_1 L_{cr} + \gamma_2 L_{ccp} + \gamma_3  L_{div} + \gamma_4 L_{ctr},
\end{equation}
and we set $\gamma_1 = \gamma_2 = \gamma_3 = \gamma_4 = 1.0$ during training.

\section{Experiments}
\label{sec:experiment}

\subsection{Experimental Setup}
\label{sec:experiment setup}

\textbf{Datasets.}
\textbf{VisDA-C} \cite{peng2017visda} is a large-scale dataset consisting of 12 classes for Synthetic-to-Real object classification. The source domain has 152k synthetic images, while the target domain has 55k real-world images. We compare the per-class top-1 accuracies and their class-wise averages. \textbf{DomainNet-126} is a subset of DomainNet \cite{peng2019moment}, comprising 4 domains: Real (R), Sketch (S), Clipart (C), and Painting (P) with 126 classes, following the setup in \cite{saito2019semi}. \textbf{PACS} \cite{li2017deeper} has 4 domains: Art-Painting (A), Cartoon (C), Photo (P), and Sketch (S) with 7 classes. For DomainNet-126 and PACS, we compare the accuracy for each domain shift and the average of all domain shifts. We will perform experiments that deal with the domain shifts within each dataset.

\noindent
\textbf{Architecture.} For a fair comparison with prior methods, we adopt the same architectures and training strategies for the source model. Specifically, we use ResNet \cite{he2016identity} as our backbones: ResNet-18 for PACS, ResNet-50 for DomainNet-126, and ResNet-101 for the VisDA-C. We add a 256-dimensional bottleneck layer, which is a fully-connected layer, followed by BatchNorm \cite{ioffe2015batch} after the backbone, and apply WeightNorm \cite{salimans2016weight} to the classifier, as done in \cite{liang2020we, litrico2023guiding, chen2022contrastive}.

\noindent
\textbf{Implementation details.} {\em For the source training}, we initialize the model with ImageNet-1K \cite{deng2009imagenet} pre-trained weights. We train the pre-trained model on the source data, the same as in \cite{chen2022contrastive}. {\em For offline SFDA}, we opt for the SGD optimizer with a momentum of 0.9 and a learning rate of 2e-4 for all datasets. We set the threshold $\alpha$ of the clean probability to 0.5, the temperature $\tau$ to 0.07, the memory bank size $M$ to 16384, and the momentum value for the EMA update to 0.999. {\em For online SFDA}, we turn on/off a soft-voting for pseudo labeling when 1024 features-probability pairs are accumulated in the memory queue. Other hyper-parameters are the same as in offline SFDA. More details in Appendix.

\begin{table*}[t]
  \centering
  \resizebox{\textwidth}{!}{
  \begin{tabular}{@{}l|c|c| c c c c c c c c c c c c |c }
    \toprule
    Method & SF & On & plane & bcycl & bus & car & horse & knife & mcycl & person & plant & sktbrd & train & truck & Avg.\\
    \midrule
    MCC(ECCV'20)\cite{jin2020minimum}               &\xmark  &\xmark
    &88.7           &80.3 &80.5 &71.5 &90.1 &93.2 &85.0 &71.6 &89.4 &73.8 &85.0 &36.9 &78.8 \\
    % STAR(CVPR’20) \cite{lu2020stochastic}           &\xmark &95.0           &84.0 &84.6 &73.0 &91.6 &91.8 &85.9 &78.4 &94.4 &84.7 &87.0 &42.2 &82.7 \\
    RWOT(CVPR’20) \cite{xu2020reliable}             &\xmark  &\xmark&95.1           &80.3 &83.7 &90.0 &92.4 &68.0 &92.5 &82.2 &87.9 &78.4 &90.4 &68.2 &84.0 \\
    % SE (ICLR'18) \cite{french2017self}              &\xmark &95.9           &87.4 &85.2 &58.6 &96.2 &95.7 &90.6 &80.0 &94.8 &90.8 &88.4 &47.9 &84.3 \\
    % CAN (CVPR '19) \cite{kang2019contrastive}       &\xmark &97.0           &87.2 &82.5 &74.3 &97.8 &96.2 &90.8 &80.7 &96.6 &96.3 &87.5 &59.9 &87.2 \\
    \midrule
    Source Only                                     &-      &- &57.2           &11.1 &42.4 &66.9 &55.0 &4.4  &81.1 &27.3 &57.9 &29.4 &86.7 &5.8  &43.8\\ 
    SHOT (ICML'20) \cite{liang2020we}               &\cmark  &\xmark&95.3           &87.5 &78.7 &55.6 &94.1 &94.2 &81.4 &80.0 &91.8 &90.7 &86.5 &59.8 &83.0\\
    % $A^{2}$Net (ICCV'21) \cite{xia2021adaptive}     &\cmark &94.0           &87.8 &85.6 &66.8 &93.7 &95.1 &85.8 &81.2 &91.6 &88.2 &86.5 &56.0 &84.3\\
    % G-SFDA (ICCV'21) \cite{yang2021generalized}     &\cmark &96.1           &88.3 &85.5 &74.1 &97.1 &95.4 &89.5 &79.4 &95.4 &92.9 &89.1 &42.6 &85.4\\
    NRC (NeurIPS’21) \cite{yang2021exploiting}      &\cmark &\xmark &96.8           &91.3 &82.4 &62.4 &96.2 &95.9 &86.1 &80.6 &94.8 &94.1 &90.4 &59.7 &85.9\\
    PS (AAAI'22) \cite{du2023generation}     &\cmark  &\xmark&95.3           &86.2 &82.3 &61.6 &93.3 &95.7 &86.7 &80.4 &91.6 &90.9 &86.0 &59.5 &84.1\\
    ProxyMix (NN'22) \cite{ding2023proxymix}     &\cmark  &\xmark&95.4           &81.7 &87.2 &79.9 &95.6 &96.8 &92.1 &85.1 &93.4 &90.3 &89.1 &42.2 &85.7\\
    SFDA-DE (CVPR'22) \cite{ding2022source}         &\cmark  &\xmark&95.3           &91.2 &77.5 &72.1 &95.7 &\textbf{97.8} &85.5 &86.1 &95.5 &93.0 &86.3 &61.6 &86.5\\
    CoWA-JMDS (ICML’22) \cite{lee2022confidence}    &\cmark  &\xmark&96.2           &89.7& 83.9& 73.8& 96.4& {97.4}& 89.3& {86.8}& 94.6& 92.1& 88.7& 53.8& 86.9\\
    $\text{AaD-SFDA}^{\dagger}$ (NeurIPS’22) \cite{yang2022attracting} &\cmark  &\xmark &97.4             &90.5 &80.8 &76.2 & {97.3} &96.1 &89.8 &82.9 &95.5 &93.0 &92.0 &{64.7} &88.0\\
    $\text{AdaContrast}^{\dagger}$ (CVPR'22) \cite{chen2022contrastive}     &\cmark  &\xmark &97.2           &83.6 &84.0 &77.2 &96.9 &94.4 &91.6 &84.9 &94.5 &93.3 &{94.1} &45.9 &86.5\\
    NRC++ (TPAMI’23) \cite{yang2021exploiting}      &\cmark  &\xmark &97.4           &\textbf{91.9} &{88.2} &{83.2} &97.3 &96.2 &90.2 &81.1 &96.3 &{94.3} &91.4 &49.6 &88.1\\
    $\text{C-SFDA}^{\dagger}$ (CVPR'23) \cite{karim2023c}              &\cmark  &\xmark&{97.6} &88.8 &86.1 &72.2 &97.2 &94.4 &92.1 &84.7 &93.0 &90.7 &93.1 &63.5 &87.8\\
    $\text{GU-SFDA}^{\dagger}$ (CVPR'23) \cite{litrico2023guiding} &\cmark  &\xmark &97.1  &{91.2} &87.6 &72.1 &96.9 &96.4 &{93.8} &86.7 &{96.3} &94.2 &91.3 &\textbf{67.0} & 89.2\\ 
    \rowcolor{Gray} CNG-SFDA                        &\cmark  &\xmark &\textbf{98.0}  &90.2 &\textbf{91.3} &\textbf{88.9} &\textbf{98.1} &96.2 &\textbf{94.7} &\textbf{88.6} &\textbf{97.7} &\textbf{97.3} &\textbf{95.3} &49.5 &\textbf{90.5}\\
    \midrule
    $\text{Tent}^{\dagger}$ (ICLR'21) \cite{wang2020tent} &\cmark  &\cmark &86.9 &57.7 &77.4 &56.8 &87.3 &62.4 &86.6 &62.9 &{71.2} &39.9 &84.8 &24.7 &66.5\\
    $\text{SHOT}^{\dagger}$ (ICML'21) \cite{liang2020we} &\cmark  &\cmark  &90.5 &77.0 & 76.2 & 47.5 & 87.9 & 62.1 & 75.9 & 74.4 &{83.3} & 47.0 & 84.2 & 41.6 & 70.6\\
    $\text{AdaContrast}^{\dagger}$ (CVPR'22) \cite{chen2022contrastive} &\cmark &\cmark &95.0 &68.0 &82.7 &{69.6} &94.3 &80.8 &90.3 &79.6 &{90.6} &69.7 &{87.6} &36.0 &78.7\\
    $\text{C-SFDA}^{\dagger}$  (CVPR'23) \cite{karim2023c}     &\cmark &\cmark  &{95.9} &{75.6} &{88.4} &68.1 &{95.4} &{86.1} &\textbf{94.5} &{82.0} &89.2 &{80.2} &87.3 &{43.8} & {82.1}\\
    $\text{TeSLA}^{\dagger}$  (CVPR'23) \cite{tomar2023tesla}     &\cmark &\cmark & {95.4} & \textbf{87.4} &{83.8} &{70.1} & {95.1} & {90.0} &{84.8} & {83.2} &{93.6} &{67.9} &85.4 & \textbf{49.3} & {82.2}\\
    \rowcolor{Gray} CNG-SFDA              
    &\cmark &\cmark &\textbf{96.7}	&{82.2}	&\textbf{89.3}	&\textbf{82.7}	&\textbf{96.9}	&\textbf{94.8}	&{94.2}	&\textbf{86.6}	&\textbf{94.9}	&\textbf{91.4}	&\textbf{89.8}	&{42.3}	&\textbf{86.8}\\
    \bottomrule
    \end{tabular}
    }
  \caption{Classification Accuracy (\%) on \textbf{VisDA-C} for Single-Source Domain Adaptation (ResNet-101 Backbone).}
  \label{tab:visdac}
\end{table*}

\begin{table}[h]
  \centering
  \begin{minipage}[t]{0.55\linewidth}
    \centering
    \resizebox{\linewidth}{!}{%
      \begin{tabular}{l | c | c | c c c c c c c | c}
        \toprule
        Method & SF & On & R $\rightarrow$ C & R $\rightarrow$ P & P $\rightarrow$ C & C $\rightarrow$ S & S $\rightarrow$ P & R $\rightarrow$ S & P $\rightarrow$ R & Avg \\
        \midrule
        MCC \cite{jin2020minimum}  &\xmark &\xmark&44.8 &65.7 &41.9 &34.9 &47.3 &35.3 &72.4 &48.9 \\
        \midrule
        Source Only                                 &-  & -    &55.5               &62.7               &53.0               &46.9               &50.1                   &46.3                &75.0              &55.6 \\
        SHOT \cite{liang2020we}           &\cmark &\xmark&67.7               &68.4               &66.9               &60.1               &66.1                   &59.9                &{80.8} &67.1 \\
        $\text{AdaContrast}^{\dagger}$  \cite{chen2022contrastive} &\cmark &\xmark &69.7               &69.0               &68.6               &58.4               &66.6                   &60.5                &80.2              &67.6 \\
        $\text{C-SFDA}^{\dagger}$  \cite{karim2023c}          &\cmark&\xmark &70.8               &{71.1}   &68.5               &{62.1}   &67.4                   &{62.7}    &80.4              &69.0 \\
        $\text{GU-SFDA}^{\dagger}$ \cite{litrico2023guiding} &\cmark &\xmark&\textbf{74.2}      &70.4               &{68.8}   &\textbf{64.0}      &{67.5} &\textbf{65.7}       &76.5              &{69.6} \\
        \rowcolor{Gray}CNG-SFDA                   &\cmark &\xmark&{73.7}  &\textbf{72.2}      &\textbf{71.7}      &60.6               &\textbf{67.8}          &64.7                &\textbf{83.2}     &\textbf{70.4}\\
        \midrule
        $\text{Tent}^{\dagger}$ \cite{wang2020tent}          &\cmark &\cmark&58.5               &65.7               &57.9               &48.5               &52.4                   &54.0                &67.0              &57.7 \\
        $\text{AdaContrast}^{\dagger}$ \cite{chen2022contrastive} & \cmark &\cmark&61.1 &66.9 &60.8 &53.4 &62.7 &\textbf{64.5} &78.9 &62.6 \\
        $\text{C-SFDA}^{\dagger}$ \cite{karim2023c}          & \cmark &\cmark&{61.6} &{67.4} &{61.3} &{55.1} &{63.2} &54.8 &{78.5} &{63.1}\\
        \rowcolor{Gray}CNG-SFDA                    & \cmark &\cmark&\textbf{67.1}	&\textbf{69.0}	&\textbf{65.6}	&\textbf{59.1}	&\textbf{66.4}	&{59.9}	&\textbf{81.5}	&\textbf{66.9}\\
        \bottomrule
      \end{tabular}%
    }
    \caption{Classification Accuracy (\%) on \textbf{DomainNet-126} for Single-Source Domain Adaptation (ResNet-50 Backbone).}
    \label{tab:domainnet}
  \end{minipage}
  \hfill
  \begin{minipage}[t]{0.4\linewidth}
    \centering
    \resizebox{\linewidth}{!}{%
      \begin{tabular}{l | c | c |  c c c | c c c c}
        \toprule
        Method & SF & On & P $\rightarrow$ A & P $\rightarrow$ C & P $\rightarrow$ S & A $\rightarrow$ P & A $\rightarrow$ C & A $\rightarrow$ S & Avg \\
        \midrule
        Source Only & - & - & 58.0 & 21.1 & 27.8 & 96.1 & 48.6 & 39.7 & 48.6 \\
        NEL \cite{ahmed2022cleaning} & \cmark &\xmark & 82.6 & {80.5} & 32.3 & {98.4} & {84.3} & 56.1 & 72.4 \\
        $\text{GU-SFDA}^{\dagger}$ \cite{litrico2023guiding} & \cmark &\xmark & {88.6} & \textbf{82.2} & \textbf{69.0} & 96.3 & \textbf{84.6} & {73.9} & \textbf{82.4} \\
        \rowcolor{Gray}CNG-SFDA & \cmark &\cmark & \textbf{92.2} & 69.1 & {66.2} & \textbf{98.8} & 83.0 & \textbf{84.8} & \textbf{82.4} \\
        \midrule
        $\text{Tent}^{\dagger}$ \cite{wang2020tent} & \cmark &\cmark & {74.3} & {62.1} & {58.8} & 96.3 & {72.6} & {71.1} & {72.5} \\
        $\text{SHOT}^{\dagger}$ \cite{liang2020we} & \cmark &\cmark & {76.1} & {59.1} & {60.8} & 96.9 & {73.1} & {73.7} & {73.2} \\
        \rowcolor{Gray}CNG-SFDA & \cmark &\cmark &\textbf{76.3}	&\textbf{64.4}	&\textbf{60.7}	&\textbf{97.1}	&\textbf{74.9}	&\textbf{73.3} &\textbf{74.4} \\
        \bottomrule
      \end{tabular}%
    }
    \caption{Classification Accuracy (\%) on \textbf{PACS} for Single-Source Domain Adaptation (ResNet-18 Backbone).}
    \label{tab:pacs}
  \end{minipage}
\end{table}

\subsection{Experimental Results}
\label{sec:experiment results}
Table. \ref{tab:visdac}, \ref{tab:domainnet}, and \ref{tab:pacs} show the performance of CNG-SFDA and the compared methods across 3 different datasets in the single-source setting. Table. \ref{tab:MSDA pacs} shows results in the multi-source setting. In each table, $\dagger$ indicates the results that we reproduced from the source code, and `SF' and `On' denote `source-free' and `online', respectively. All reported results are average accuracies from three seeds.

\noindent
\textbf{Single-Source DA.}
Table. \ref{tab:visdac}, \ref{tab:domainnet}, and \ref{tab:pacs} show the classification accuracy of CNG-SFDA and previous methods for UDA/SFDA in an offline setting. In Table. \ref{tab:visdac}, CNG-SFDA outperforms UDA methods such as CAN \cite{kang2019contrastive} without access to the source data. In the more challenging SFDA, we surpass SOTA \cite{litrico2023guiding} by a 1.3$\%$ margin on per-class average accuracy  in VisDA-C (Table. \ref{tab:visdac}) and 0.8$\%$ margin on DomainNet-126 (Table. \ref{tab:domainnet}). Moreover, in PACS (Table. \ref{tab:pacs}), we achieve competitive performance compared to SOTA. Unlike self-training methods (AdaContrast \cite{chen2022contrastive}, GU-SFDA \cite{litrico2023guiding}) using nearest features for pseudo-labels, CNG-SFDA leverages clean probabilities within clusters, proving more effective.

% The self-training methods, such as AdaContrast \cite{chen2022contrastive} and GU-SFDA \cite{litrico2023guiding}, utilize the nearest features to generate reliable pseudo-labels without considering cluster structure. We believe that CNG-SFDA is more effective than these methods since it addresses the pseudo-labels based on the clean probabilities within the cluster.

\noindent
\textbf{Multi-Source DA.}
Table. \ref{tab:MSDA pacs} shows the performance of Multi-Source Unsupervised Domain Adaptation (MSUDA) on PACS.
In MSUDA, each domain is treated as the target, while the source model is trained by aggregating all other domains without domain labels.
In source-free MSUDA, CNG-SFDA achieves the highest accuracy by training pseudo-labels for the target with different strategies depending on whether the data is in clean or noisy regions, whereas SHOT++ \cite{liang2021source} produces pseudo-labels using the centroid of the nearest cluster.
\begin{table}[t]
\null
\centering
\begin{minipage}[t]{0.48\linewidth}
\centering
\resizebox{\linewidth}{!}{%
\begin{tabular}{l | c | c | c c c c | c }
\toprule
Method & SF & On & A & C &  P & S & Avg \\
\midrule
SIB \cite{hu2020empirical} & \xmark & \xmark & 88.9 & 89.0 & 98.3 & 82.2& 89.6 \\
T-SVDNET  \cite{li2021t} & \xmark & \xmark& 90.4 & 90.6 & 98.5 & 85.4 & 91.2 \\
iMSDA\cite{kong2022partial} & \xmark & \xmark& 93.7 & 92.4 & 98.4 & 89.2 & 93.4 \\
\midrule
Source Only & - & - & 77.9 & 72.8 & 95.7 & 63.5 & 77.5 \\
$\text{SHOT}^{\dagger}$ \cite{liang2020we} & \cmark& \cmark & 90.7 & 88.1 & 98.5 & 75.4 & 88.2 \\
$\text{SHOT++}^{\dagger}$ \cite{liang2021source} & \cmark & \cmark& {92.3} & {89.7} & \textbf{98.8} & {75.5} & {89.1} \\
\rowcolor{Gray} CNG-TTA & \cmark & \cmark & \textbf{93.7} & \textbf{93.3} & {98.7} & \textbf{87.8} & \textbf{93.4} \\
\bottomrule
\end{tabular}
}
\hfill
\caption{Classification Accuracy (\%) on PACS for Multi-Source Domain Adaptation (ResNet-18 Backbone).}
\label{tab:MSDA pacs}
\end{minipage}%
\begin{minipage}[t]{0.48\linewidth}
\centering
\resizebox{\linewidth}{!}{%
\begin{tabular}{c | c c c c c | c c c}
\toprule
PL & $L_{cr}$ & $L_{ccp}$ & $L_{div}$ &  $L_{inst}$ & $L_{prt}$ & VisDA-C & DomainNet & PACS \\
\midrule
\cmark & \xmark & \xmark & \xmark & \xmark  & \xmark & 84.9 & 58.5 & 75.7 \\
\cmark & \cmark & \xmark & \xmark &  \xmark & \xmark & 88.0 & 65.8 & 78.8 \\
\cmark & \cmark & \cmark & \xmark & \xmark  & \xmark & 89.6 & 67.4 & 79.7 \\
\cmark & \cmark & \cmark & \cmark & \xmark  & \xmark & 89.0 & 67.7 & 80.6 \\
\cmark & \cmark & \cmark & \cmark & \cmark & \xmark & 90.1 & 69.7 & 81.1 \\
\rowcolor{Gray}\cmark & \cmark & \cmark & \cmark & \cmark & \cmark & \textbf{90.5} & \textbf{70.4} & \textbf{82.4} \\
\bottomrule
\end{tabular}
}
\caption{The effectiveness of each component across \textbf{all datasets} is validated by classification accuracy (\%).}
\label{tab:component_abaltion}
\end{minipage}
\end{table}

\noindent
\textbf{Online Source-Free DA.}
In Table. \ref{tab:visdac}, \ref{tab:domainnet}, \ref{tab:pacs}, and \ref{tab:MSDA pacs}, `online' results show the classification accuracy of CNG-SFDA and the compared methods for online SFDA. In Table. \ref{tab:visdac} and \ref{tab:domainnet}, CNG-SFDA significantly outperforms SOTA(4.6$\%$ and 3.8$\%$ improvements in VisDA-C and DomainNet-126). C-SFDA \cite{karim2023c} may fail to capture all the knowledge from the target domain, particularly from the noisy data, due to its selective approach based on curriculum learning. However, CNG-SFDA utilizes the entire target domain's knowledge by training effectively all samples. Plus, Table. \ref{tab:pacs}, \ref{tab:MSDA pacs} show the effectiveness of CNG-SFDA on PACS for online SFDA. 

\begin{figure}[t]
    \centering
    \includegraphics[width=.8\linewidth]{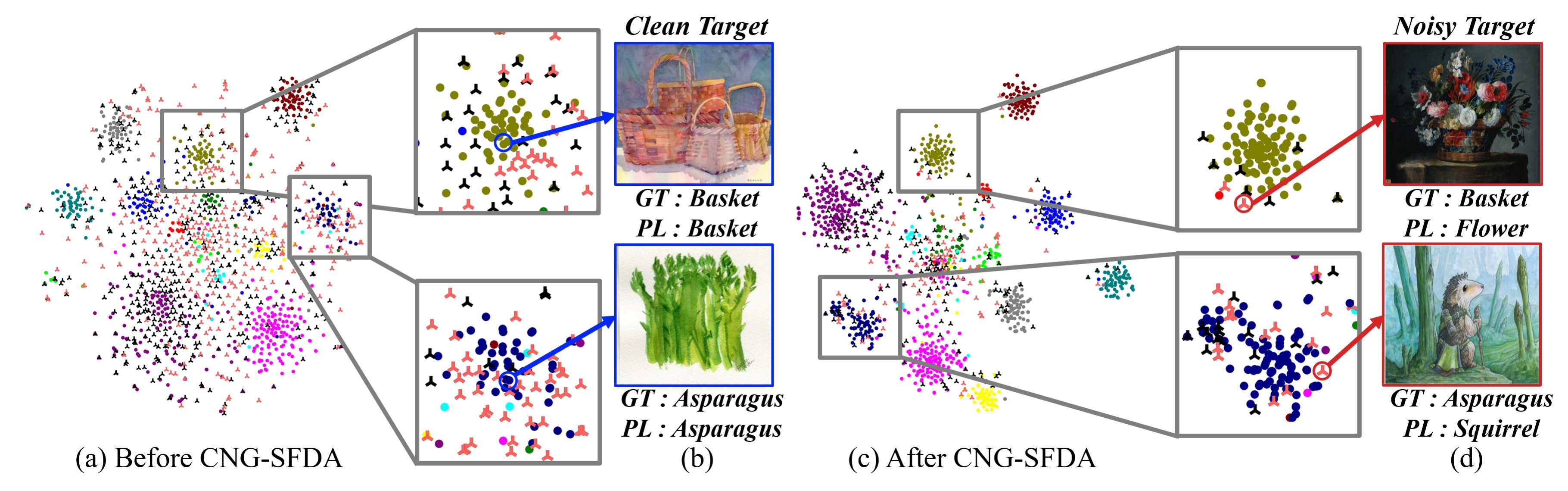}
    \caption{Feature distribution of clean and noisy samples in target domain. Here, `clean' and `noisy' refer to cases where the Ground-Truth (GT) and the Pseudo-Label (PL) match and mismatch. Circle and triangular denote clean and noisy samples predicted by CNG-SFDA. For clean samples (circle), each color represent the GT class. For noisy samples (triangular), pink and black triangular represent cases where CNG-SFDA correctly (i.e., Prediction: Noisy, GT: Noisy) and incorrectly (i.e., Prediction: Noisy, GT: Clean) identifies noisy data.}
    \label{fig:clean_noisy_fig}
\end{figure}

\subsection{Analysis and Discussion}
\label{sec:analysis and discussion}

\noindent
\textbf{Ablation study.}
Table. \ref{tab:component_abaltion} shows the effectiveness of each component on all datasets. The `PL' method as a baseline indicates the method generating the pseudo-labels from nearest features in Eq. \ref{eqation:1}. By applying `$L_{cr}$' in Sec. \ref{sec:section-b,c}, we boost the performance by 3.1$\%$, 7.3$\%$, and 3.1$\%$ on each domain. These results confirm that `$L_{cr}$' learning only the samples with clean labels is more effective than `PL' using all data for training. Plus, incorporating the cluster compactness loss `$L_{ccp}$' in Sec. \ref{sec:section-b,c} and contrastive losses `$L_{inst}$', `$L_{prt}$' in Sec. \ref{sec:section-d} consistently contributes to performance improvement on all datasets.

\begin{table}[t]
  \begin{minipage}[t]{0.48\linewidth}
        \centering
        \null
        \resizebox{\linewidth}{!}{%
            \begin{tabular}{l | c c c c | c }
                \toprule
                Method & \rotatebox[origin=c]{70}{real $\rightarrow$} & \rotatebox[origin=c]{70}{clipart $\rightarrow$} & \rotatebox[origin=c]{70}{painting $\rightarrow$} & \rotatebox[origin=c]{70}{sketch $\rightarrow$}& Avg. \\
                \midrule
                Source only  &45.3 &49.3 &41.7 &44.8  &45.3  \\
                % Tent (ICLR'21) &42.4 &44.2 &37.2 &37.5 &40.3 \\ 
                $\text{AdaContrast}^{\dagger}$ (CVPR'22) & 38.3 &38.8&33.7 & \textbf{32.7}&35.9 \\
                $\text{CoTTA}$ (CVPR'22)  &43.4 &43.0 &36.4 &36.3 &39.8 \\
                $\text{SAR}^{\dagger}$ (ICLR'23) &42.8 &44.3 &37.7 &37.7 & 40.6\\
                $\text{RoTTA}^{\dagger}$ (CVPR'23) &43.2 &44.0 &38.6 &37.6 &40.8 \\
                $\text{RMT}^{\dagger}$ (CVPR'23) &\textbf{37.6} &\textbf{38.6} &33.3 &33.3 &35.7 \\
                \rowcolor{gray!20} CNG-SFDA &37.7 &39.2 &\textbf{31.5} &33.7 &\textbf{35.5} \\
                \bottomrule
            \end{tabular}
        }
        \captionof{table}{Classification Error Rate (\%) on DomainNet-126 for Continual TTA.}
        \label{tab:continual_TTA}
    \end{minipage}
    \begin{minipage}[t]{0.48\linewidth}
    \centering
    \null
        \resizebox{\linewidth}{!}{%
        \begin{tabular}{l | c c c}
          \toprule
          BackBone & ResNet18 & ResNet50 & ViT-B/16 \\
          \midrule
          ERM \cite{Vapnik1998} & 82.1 & 84.6 & 87.1 \\
          % RSC (ECCV'20)\cite{huang2020self} & 85.2 & 87.8 & - \\
          DGCM (ICML'21)\cite{mahajan2021domain} & 85.5 & 87.5 & - \\
          FACT (CVPR'21) \cite{xu2021fourier} &84.5 & 88.2 & - \\
          T3A (NeurIPS'21) \cite{xu2021fourier} &81.7 & 84.5 & 86.0 \\
          AdaNPC (ICML'23) \cite{zhang2023adanpc} & 83.1 & 85.7 & 88.7 \\
          $\text{TSD}^{\dagger}$ (CVPR'23)\cite{wang2023feature}  & \textbf{87.8} & {89.4} & {90.2}\\
          \rowcolor{Gray}CNG-SFDA (Ours) & {86.5} & \textbf{90.4} & \textbf{91.7} \\
          \bottomrule
        \end{tabular}%
        }
        \captionof{table}{Classification Accuracy (\%) on \textbf{PACS} with different backbone architecture for Multi-Source Unsupervised Domain Adaptation.}
        \label{tab:compare_backbone}
  \end{minipage}
\end{table}

\begin{figure}[!t]
    \centering
    \includegraphics[width=.95\textwidth]{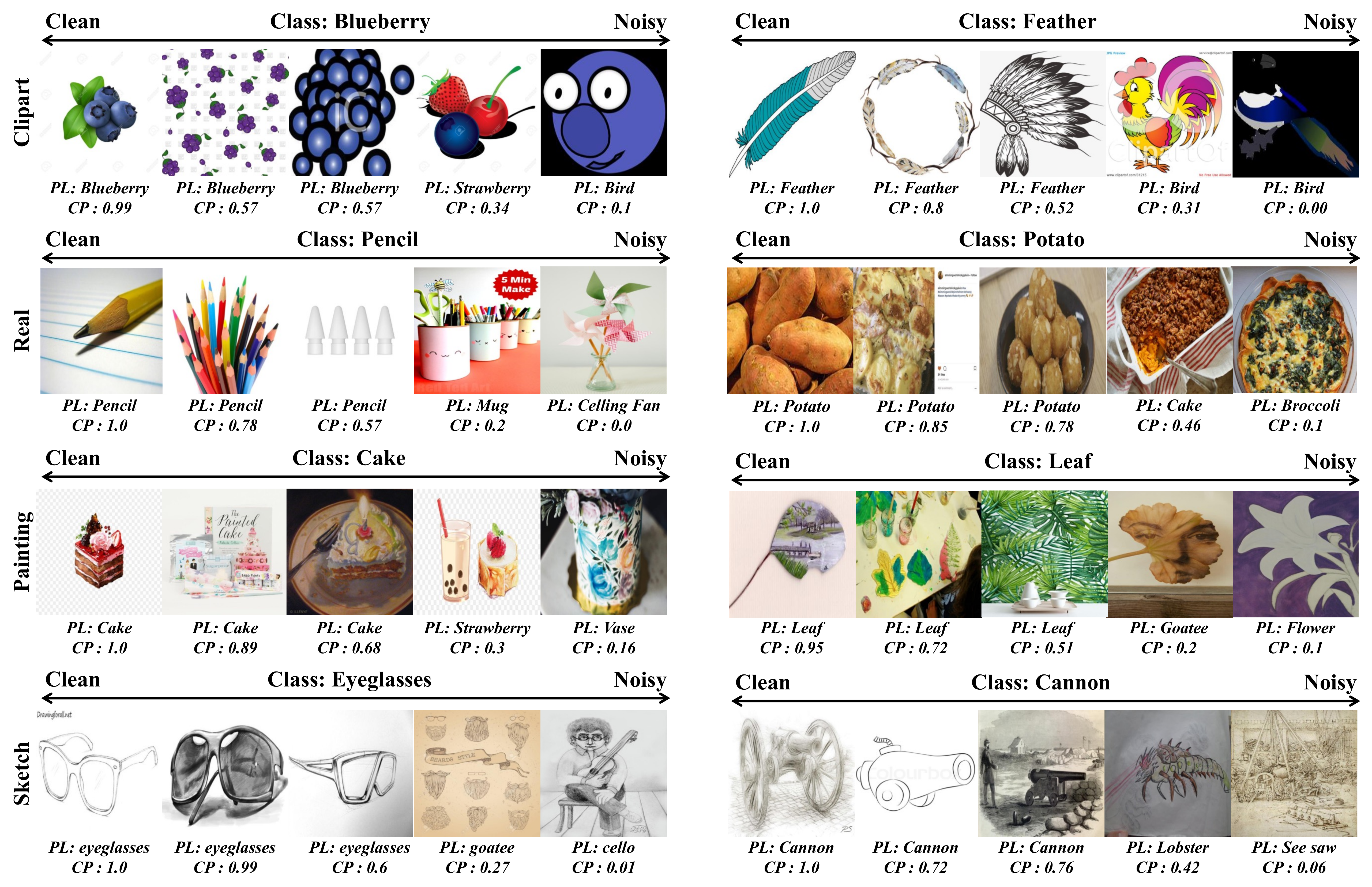}
    \caption{Qualitative evaluation of samples in predicted clean and noisy regions by CNG-SFDA at each target domain (each row). PL and CP are Pseudo-Label and Clean Probability estimated by CNG-SFDA, respectively.
    }
    \label{fig:Clean and Noise Example}
\end{figure}

\pgfplotsset{compat=1.18}
\pgfplotsset{every tick label/.append style={font=\normalsize}}
\pgfplotsset{every x tick label/.append style={font=\normalsize, yshift=0.5ex}}
\pgfplotsset{every y tick label/.append style={font=\normalsize, xshift=0.5ex}}
\pgfplotsset{every y tick label/.append style={font=\normalsize, xshift=0.5ex}}
\pgfplotsset{every axis/.append style={
label style={font=\normalsize},tick label style={font=\normalsize}}
}
\pgfplotsset{every axis title/.append style={at={(0.5,-0.4)}}}

\tikzset{every mark/.append style={scale=1.2}}
\tikzset{every path/.style={line width=0.07 cm}}

% \pgfplotsset{legend image code/.code={%
% \draw[#1] (0cm,0cm) rectangle (0.6cm,0.1cm);}
% }

\pgfplotstableread{sensitivity_clean_prob.csv}{\datatable}
\pgfplotstableread{sensitivity_k_near_contrast.csv}{\kdatatable}
\pgfplotstableread{online_start_length.csv}{\onlinetable}
\usepgfplotslibrary{groupplots}

\begin{figure}[t]
    \begin{minipage}[t]{0.48\linewidth}
        \null
        \centering
        \resizebox{\linewidth}{!}{%
            \begin{tikzpicture}
    
            \begin{groupplot}[group style={group size=3 by 1,
                                },
                legend pos=south east
            ]
            \nextgroupplot[
                axis line style = thin,
                legend entries={Accuracy, Average},
                legend style={fill=gray!15,draw=white},
                legend image post style={mark=*},
                axis background/.style={fill=gray!15},
                grid=major, 
                grid style={white},
                axis lines = left, 
                xlabel=threshold,
                ylabel=score,
                ymin=87,ymax=91,
                xmin=-0.04,xmax=1
                ]
                % \addlegendimage{no markers, line}
                % \addlegendimage{no markers}
                        
                \addplot[
                mark=*, strongblue
                ] table [
                  x=time,
                  x expr=\thisrow{time},
                  y=Accuracy,
                ]{\datatable};
                % \addlegendentry{accuracy}
                
                \addplot[
                mark=*, vividorange
                ] table [
                  x=time,
                  x expr=\thisrow{time},
                  y=Average,
                ]{\datatable};
                
            \nextgroupplot[
                axis line style = thin,
                axis background/.style={fill=gray!15},
                grid=major, 
                grid style={white},
                axis lines = left, 
                legend entries={Accuracy, Average},
                legend style={fill=gray!15,draw=white},
                legend image post style={mark=*},
                xlabel=N,
                % ylabel=$Score$,
                ymin=88,ymax=91,
                xmin=-2,xmax=45
                ]
                
                \addplot[
                mark=*, strongblue
                ] table [
                  x=time,
                  x expr=\thisrow{time},
                  y=Accuracy,
                ]{\kdatatable};
                % \addlegendentry{accuracy}
                
                \addplot[
                mark=*, vividorange
                ] table [
                  x=time,
                  x expr=\thisrow{time},
                  y=Average,
                ]{\kdatatable};
        
            \nextgroupplot[
                axis line style = thin,
                axis background/.style={fill=gray!15},
                grid=major, 
                grid style={white},
                axis lines = left, 
                legend entries={Accuracy, Average},
                legend style={fill=gray!15,draw=white},
                legend image post style={mark=*,},
                xlabel=Memory Length,
                ymin=84,ymax=88,
                xmin=4,xmax=10.5,
                xticklabel={$2^{\pgfmathprintnumber{\tick}}$}
                ]
                
                \addplot[
                mark=*, strongblue
                ] table [
                  x=time,
                  x expr=\thisrow{time},
                  y=Accuracy,
                ]{\onlinetable};
                
                \addplot[
                mark=*, vividorange
                ] table [
                  x=time,
                  x expr=\thisrow{time},
                  y=Average,
                ]{\onlinetable};
                
            \end{groupplot}
        \end{tikzpicture}
        }
        \caption{Robustness analysis of different hyper-parameters.}
        \label{fig:sensitivity}
    \end{minipage}%
    \hfill
    \begin{minipage}[t]{0.45\linewidth}
        % \centering
        \null
        \resizebox{\linewidth}{!}{%
        \begin{tabular}{l | c c c}
          \toprule
          Method & VisDa-C & DomainNet & PACS \\
          \midrule
          w/o weighting & 89.7 & 69.2 & 80.7 \\
          w/ linear weighting & 90.0 & 70.2 & 82.1\\
          w/ exponential weighting & \textbf{90.5} & \textbf{70.4} & \textbf{82.4} \\
          \bottomrule
        \end{tabular}%
        }
        \captionof{table}{Classification Accuracy (\%) on \textbf{all dataset} comparing linear vs exponential weighting in Eq. \ref{eqation:13}.}
        \label{tab:compare_linear_exp}
    \end{minipage}
\end{figure}

\noindent
\textbf{Effect on clean and noisy region.} 
Fig. \ref{fig:clean_noisy_fig} shows the effectiveness of distinguishing clean and noisy data within clusters. From the gray boxes in (a) and (c) of Fig. \ref{fig:clean_noisy_fig}, we observed that applying CNG-SFDA resulted in a reduction of noisy samples and clustering around clean samples. Additionally, as shown in (a), CNG-SFDA correctly identified noisy samples early in training, effectively utilizing them with clean samples. Interestingly, as seen in (d), samples correctly predicted to have noisy pseudo-labels were indeed difficult samples to classify.

\noindent
\textbf{Capability of Continual Test-Time Adaptation.}
We also conducted experiments on DomainNet-126 in continual Test-time adaptation settings \cite{wang2022continual}. 
Continual TTA is particularly useful for scenarios where the target data distribution is not known in advance or is constantly changing.
Consequently, the source model is dynamically adapted to a sequence of test domains in an online manner. 
For DomainNet-126, we randomly generated four different domain sequences, ensuring that each domain serves as the source domain once. 
Table. \ref{tab:continual_TTA} reveals that CNG-SFDA surpasses SOTA \cite{dobler2023robust} for the continual TTA, proving that its robustness in dynamic scenarios where target distributions are changing.

\noindent
\textbf{Qualitative evaluation on distinguishing clean and noisy regions.}
Fig. \ref{fig:Clean and Noise Example} showcases samples with clean and noisy pseudo-labels predicted by CNG-SFDA, along with the PL and CP of each sample. 
We observed that noisy images tend to have lower CP values, indicating difficulty in classification. 
These noisy images often contain objects outside the class or are challenging to perceptually classify. 
On the other hand, clean images mostly comprise easy samples containing only objects corresponding to the ground truth classes. This qualitative analysis confirms the reliability of CNG-SFDA in identifying clean and noisy images.

\noindent
\textbf{Scalability on different models.}
In Table. \ref{tab:compare_backbone}, we validate our method using different backbones to ensure it is the model-agnostic method. CNG-SFDA outperforms better than the previous methods on PACS for MSUDA. Through these results, we confirm that our method works well with different backbones.

\noindent
\textbf{Robustness of hyper-parameters.}
Fig. \ref{fig:sensitivity} shows the robustness of CNG-SFDA versus the clean probability threshold $\alpha$ and $K$ in Eq. \ref{eqation:17}. In addition, we analyze the sensitivity of the memory queue length, which is the criterion for turning on-off soft voting for pseudo-labeling for online SFDA. These results demonstrate the robustness of CNG-SFDA to the hyper-parameters.

\noindent
\textbf{Linear vs Exponential weight.}
Table. \ref{tab:compare_linear_exp} presents the effectiveness of the exponential weighting in Eq. \ref{eqation:13}, comparing linear one and without one. CNG-SFDA outperforms previous SFDA methods, even with linear weighting.

\section{Conclusion}
\label{sec:conclusion}
We introduced CNG-SFDA, a novel online-offline SFDA approach in image classification. To ensure the comprehensive transfer of target domain knowledge to the source model, effective learning of clean and noisy pseudo-labels from the target domain is crucial. To address this, we propose to distinguish clean and noisy regions at the cluster level. Then, we apply different training strategies to each region by considering their respective characteristics. Our experiments and analysis prove the effectiveness of CNG-SFDA compared to other approaches.

\noindent
\textbf{\ackname} This research was conducted with resources and endless support from VUNO Inc, and Won Hwa Kim was supported by Graduate School of AI at POSTECH (IITP-2019-0-01906).

%
% ---- Bibliography ----
%
% BibTeX users should specify bibliography style 'splncs04'.
% References will then be sorted and formatted in the correct style.
%
% \bibliographystyle{splncs04}
% \bibliography{mybibliography}
%
\bibliographystyle{splncs04}
\bibliography{main}
\end{document}